# Physical Logic Enhanced Network for Small-Sample Bi-Layer Metallic Tubes Bending Springback Prediction

Chang Sun[1], Zili Wang[1,2(✉)], Shuyou Zhang[1,2], Le Wang[1], Jianrong Tan[1,2]

1. State Key Laboratory of Fluid Power and Mechatronic Systems, Zhejiang University, Hangzhou, 310027, China.
2. Engineering Research Center for Design Engineering and Digital Twin of Zhejiang Province, Zhejiang University, Hangzhou, 310027, China)

e-mail: ziliwang@zju.edu.cn.

**Abstract.** Bi-layer metallic tube (BMT) plays an extremely crucial role in engineering applications, with rotary draw bending (RDB) the high-precision bending processing can be achieved, however, the product will further springback. Due to the complex structure of BMT and the high cost of dataset acquisition, the existing methods based on mechanism research and machine learning cannot meet the engineering requirements of springback prediction. Based on the preliminary mechanism analysis, a physical logic enhanced network (PE-NET) is proposed. The architecture includes ES-NET which equivalent the BMT to the single-layer tube, and SP-NET for the final prediction of springback with sufficient single-layer tube samples. Specifically, in the first stage, with the theory-driven pre-exploration and the data-driven pretraining, the ES-NET and SP-NET are constructed, respectively. In the second stage, under the physical logic, the PE-NET is assembled by ES-NET and SP-NET and then fine-tuned with the small sample BMT dataset and composite loss function. The validity and stability of the proposed method are verified by the FE simulation dataset, the small-sample dataset BMT springback angle prediction is achieved, and the method potential in interpretability and engineering applications are demonstrated.

**Keywords:** physical logic enhanced network, mechanism analysis, small-sample BMT dataset, composite loss function.

## 1 Introduction

Bent-tube is an important component for transporting gas, liquid, and even load-bearing, which plays an extremely crucial role in industrial production and application. Rotary draw bending (RDB) is the primary method for bent-tube processing, which has the advantages of high precision and well flexibility. After the bending processing, the elastic deformation recovers with the removal of the mold's constraints, which is normally called the springback phenomenon. The springback will significantly affect the accuracy of the product, making a margin necessarily to be given for compensation in the processing plan with an accurate springback prediction in advance.

Through the mechanism analysis of the bending process, the single-layer tube springback can be predicted from the angles of the bending state [1], prestress field [2], etc. With the improvement of the industry's requirements, the bi-layer metallic tube (BMT) have gradually been widely applied with advantages such as corrosion protection, wear and impact resistance, thermal and electric insulation. The material properties and interlayer coupling of BMT are incredibly complex [3], which greatly increases the difficulty of springback prediction. Therefore, different from the well-established research of the single-layer metallic tube processing, the existing springback research of the BMT mainly focuses on the influence of individual factors [4, 5]. Its processing deformation mechanism still urgently needs to be further explored.

Although it is difficult for BMT to implement the forming theory of classical single-layer tubes directly, there is a remarkable similarity in bending deformation between them. In fact, it is feasible to equivalent the bi-layer materials as one of them for simplifying, which has been verified in beams and slabs [6, 7]. Nevertheless, due to the unique cross-sectional properties, the errors in the BMT equivalence make it difficult to be used directly.

Through the unique structure and learning method, the neural network can reach the effect of a nonlinear function, and have the ability to accomplish the springback prediction based on the batch FE simulation data [8, 9]. However, when it comes to a problem with the small-sample dataset, the network can be challenging to reach an acceptable accuracy.

From the perspective of the scientific paradigm, the machine learning algorithms and the physical models are driven by data and theory, respectively. The theoretical approach is based on strict logical relationships with strong interpretability and generalization performance. Meanwhile, machine learning methods can realize the reflection of the internal relationship from dataset dimensions that are difficult to observe. Therefore, the combination of them has received more and more attention. In fact, with the guidance of physical knowledge, machine learning has been shown to benefit from plausible physically-based relationships in research and applications [10, 11]. Such methods have been initially applied and demonstrated their effectiveness in disease prediction [12, 10], geological detection [13], natural language processing [14], mechanical analysis [15], etc.

As mentioned above, the deformation mechanism of the BMT is exceptionally complicated. At the same time, since the high cost of BMT and the complex processing requirements, the scale of the dataset is limited, which makes it difficult for the network to achieve acceptable accuracy with the data-driven approach alone. However, combined with the preliminary physical analysis, the network can learn start on the basis of physical logic and existing knowledge, which can improve the training efficiency, reduce the probability of overfitting, and improve the accuracy of the network.

Therefore, the physical logic enhanced network (PE-NET) is proposed to solve the problems of the insufficient mechanism analysis and limited datasets of BMT. PE-NET includes an equivalent section network (ES-NET) and a single-layer prediction network (SP-NET). ES-NET is used to map the equivalent shape parameters of BMT to single-layer tube, while SP-NET carries the knowledge of single-layer bending deformation for final prediction. The two are pre-trained under the data-driven and theory-driven, respectively, and then combined as PE-NET under the physical logic. After that, the parameters of PE-NET will be rationally constrained by the loss function combination of mechanism equation, and fine-tuned by the small BMT dataset.

The main contributions of this paper are as follows: (1) The equivalent section method is applied to the mechanism analysis of BMT bending. (2) A collaborative architecture of theory-driven and data-driven together is constructed based on physical logic. (3) The BMT springback angle prediction is realized by the proposed PE-NET with small-sample dataset and mechanism analysis.

# 2 Background knowledge

## 2.1 RDB processing of BMT

Springback is caused by residual stress after RDB processing which can be affected by multiple factors. As shown in Fig. 1, during the processing, the tube blank is deformed under the constraint and movement of molds. Specifically, under the boosting of the pressure die and the clamping of the clamp die, the tube blank is bent and rotated around the bending die, while the wiper die prevents wrinkling defect. The radius $R_B$ and rotation angle $\alpha_B$ of the bending die directly determine the product shape. At the same time, the initial location $L_P$ and boost velocity $v_B$ of the pressure die, the processing velocity $\omega_B$, the gap $G_i$ and friction $f_i$ between the tube blank and the molds also have a significant influence on the generation and distribution of residual stress.

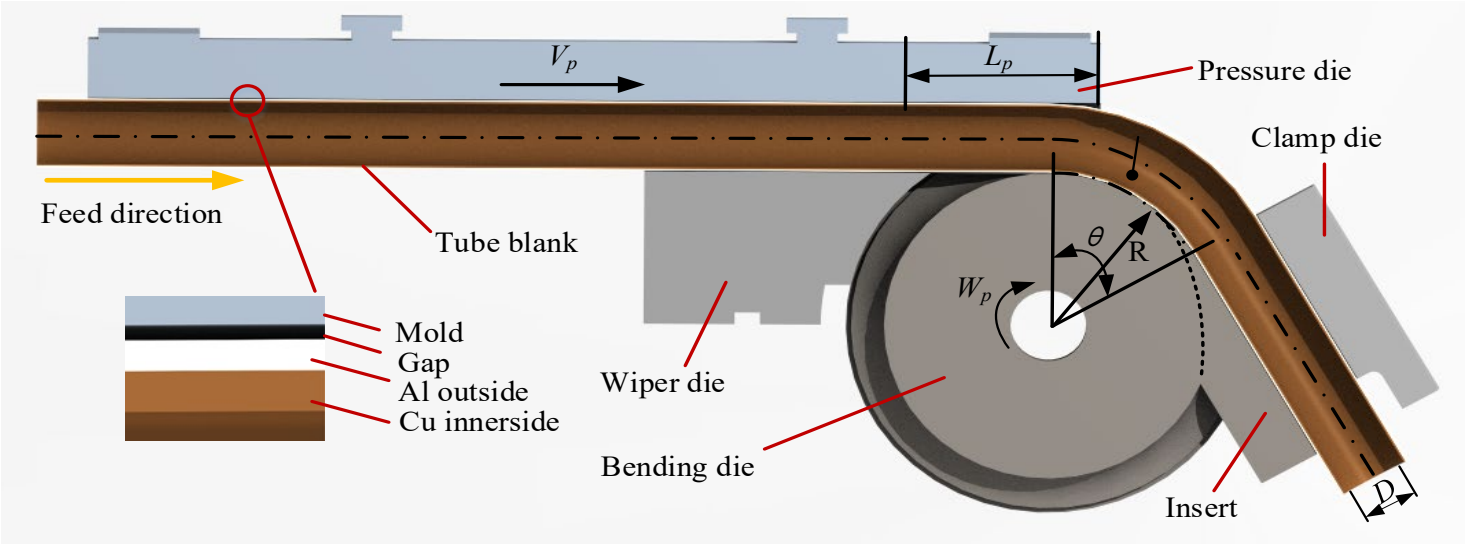

**Fig. 1.** RDB processing of BMT

In addition to the above processing parameters, the shape parameters are also major factor of springback to be considered. The wall thickness $T$ and diameter $D_o$ of the tube blank directly determine the bending properties and springback angle. For BMT, the thickness ratio $T_r$ is also one of the decisive factors.

## 2.2 Equivalent section theory

The springback can be regarded as the recovery of the elastic deformation caused by a reverse moment $M$ from the internal processing stress of the bent-tube before springback. Therefore, the equivalent characteristics of the material in the elastic deformation stage will be mainly analyzed.

The mechanism of tube bending is always based on the sheet bending theory. As shown in Fig. 2, take a micro-element on the section of the laminated beam. In order to simplify the analysis, the bending process satisfies the assumption of plane section and unidirectional force, and the neutral layer does not shift. For materials such as aluminum and alloy steel, the linearly strengthened elastic-plastic material model can be adopted. Assuming that there is no axial force during the processing, i.e., the pure bending condition, and normal stress can be regarded as zero. Combine with the stress distribution in the elastic deformation stage, we have

$$F_N = \int_{A_k} \sigma(x,y,z)\, dA = \sum_k^n \int_{A_k} E_k \frac{y}{\rho} dA = \frac{1}{\rho} \sum_k^n E_k S_{Zk} = 0 \tag{1}$$

$$M_z = \int_{A_k} \sigma(x,y,z)\, y dA = \sum_k^n \int_{A_k} E_k \frac{y^2}{\rho} dA = \frac{1}{\rho} \sum_k^n E_k I_{Zk} \tag{2}$$

Where for the $k$-th material, $S_{Zk}$ is the area moment (static moment), $I_{Zk}$ is the moments of inertia, $E_k$ is the distribution of the elastic modulus, and $\rho$ is the distance from the neutral axis.

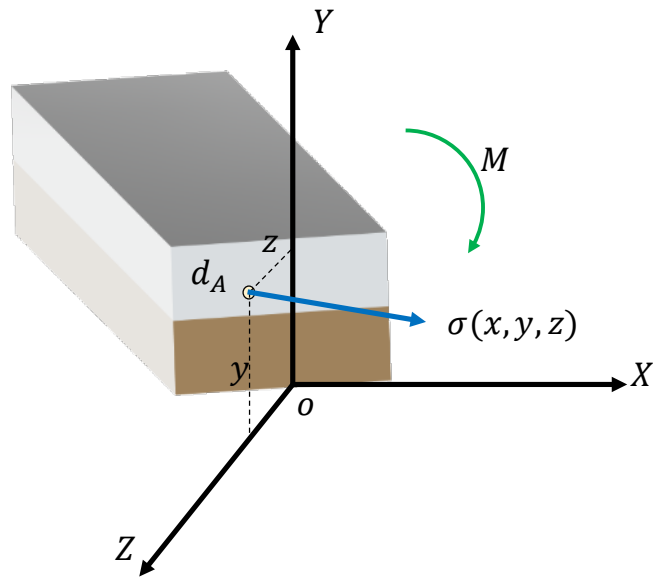

**Fig. 2.** Bending analysis of laminated beams

Then we have

$$\sigma_i = \frac{-E_i M y}{\sum_{k=1}^n E_k I_{Zk}} = \lambda_i \frac{My}{I_{Z0}} \quad (i = 1,2, \dots n) \tag{3}$$

$$I_{z0} = \sum_{k=1}^n \lambda_k I_{Zk} \tag{4}$$

$$S_{Z0} = \sum_{k=1}^n \lambda_k S_{Zk} = 0 \tag{5}$$

Where $\lambda_i = E_i / E_m$ $(i = 1,2, \dots n)$ is the ratio of the elastic modulus, $E_m$ is the reference elastic modulus, $S_{Z0}$ is the total area moment of each equivalent section, and $I_{z0}$ is the total moments of inertia of each equivalent section. For an integral region, $1/\rho$ and $E_k$ are both constants.

As shown in Eq. (4) and Eq. (5), the thickness of each material region will be scaled down by $\lambda_k$, the width and the circumferential centroid position will remain unchanged, and the multi-material section will be transformed into the specified material section. $I_{Z0}$ can be obtained by the moment of inertia of the transformed section to the neutral axis, and the equivalent single-layer tube radius and thickness can be further calculated.

# 3 Methodology

## 3.1 Proposed prediction architecture

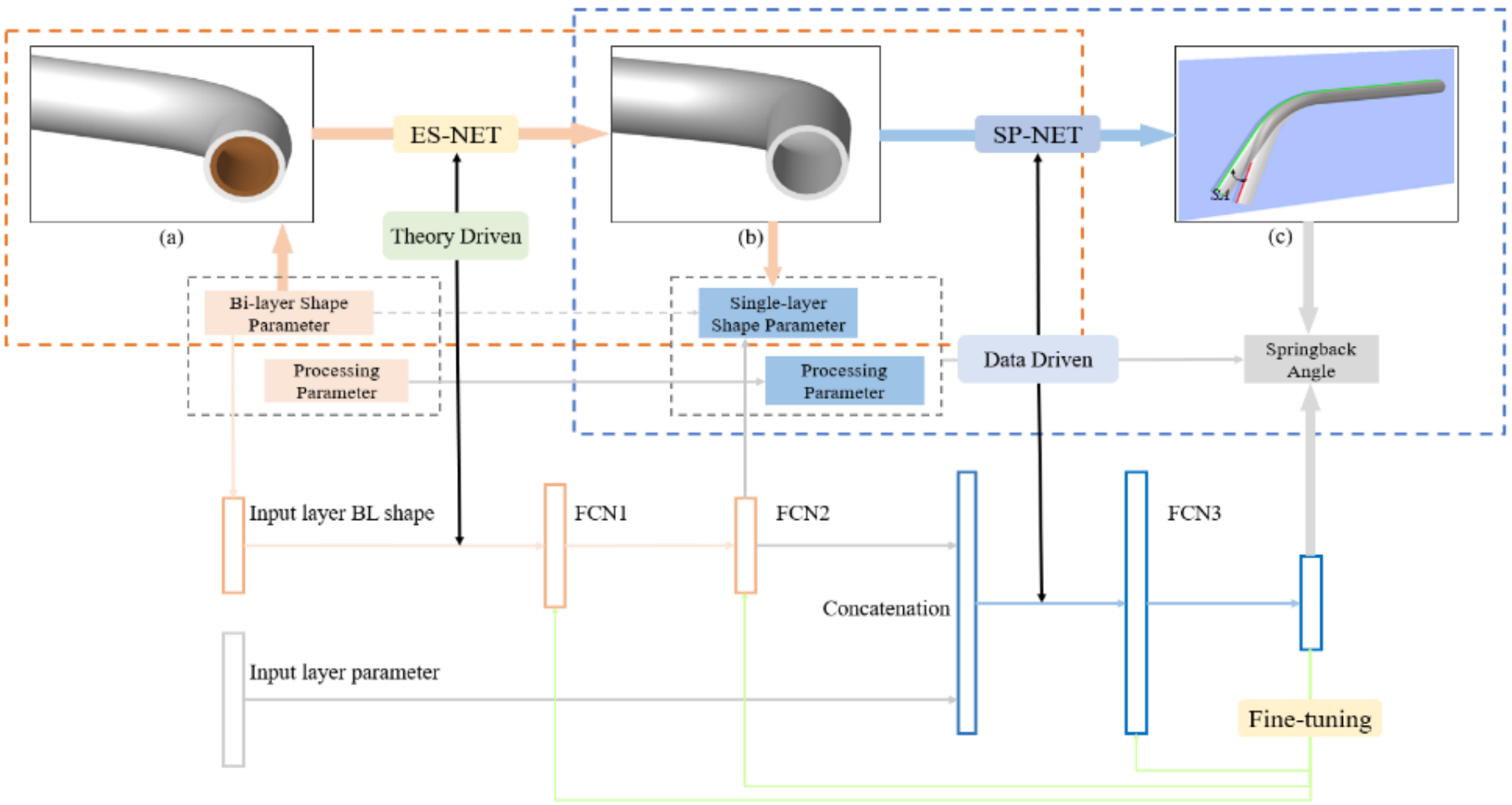

**Fig. 3.** The architecture of the PE-NET: (a) is the shape parameter of the BMT; (b) is the shape parameter of the single-layer tube; (c) is the predicted springback.

The parameter update of the traditional neural network relies on the observation of the prediction value and the label, which the parameters are able to explore in an unrestricted region. However, the small-sample of the bi-layer dataset cannot guarantee that the network parameters reach an acceptable result. Realizing domain transfer based on theory analysis, and using sufficient close domain knowledge can make up for the inferior of the small-sample dataset. On this basis, the network can be partitioned into functional modules according to physical logic, which provides very strong theoretical logic and constraints on top of the observational ones [11].

Pre-exploration of the network parameter domain and applying the guidance to the network parameters in further training is an effective way to improve training efficiency and prevent overfitting [16]. In terms of results, both theory-driven and multi-objective optimization-based parameter preselection [17] are common methods to improve network accuracy. The results of multi-objective-based methods are random, lack physical logic, and rely heavily on prior knowledge. However, theory-driven network training can achieve parameter domain pre-exploration without effective prior knowledge, and is well applied to small-sample prediction.

The proposed PE-NET architecture is shown in Fig. 3. The traditional neural network for the mapping of the parameters to the springback often adopts the data-driven method with path Fig. 3(a) directly to Fig. 3(c). Due to the similar bending mechanism of bi-layer and single-layer materials, the close domain knowledge can be used. The PE-NET consists of two modules, namely the ES-NET which is for the equivalent mapping from the BMT to the single-layer tube, and the SP-NET which is for the final springback prediction with the single-layer tube knowledge. Specifically, at the first stage, the parameters of the ES-NET are pre-explored based on preliminary mechanism analysis, realizing the Fig. 3(a) to Fig. 3(b), and the parameters of SP-NET are pre-trained on the low-cost single-layer tube dataset, realizing Fig. 3(b) to Fig. 3(c). In the second stage, the PE-NET is constructed by ES-NET and SP-NET based on the physical logic. Since the discrepancy of different material datasets and the error of mechanism analysis, the fine-tuning of the PE-NET is implemented. The loss function of PE-NET in the second stage is based on two parts, i.e., the rationality constraints based on the mechanism equation, and backpropagation from the small-sample dataset. The detail of their cooperation will be introduced in Sect. 3.3.

As shown in Fig. 3, benefiting from the sufficient dataset and the ability of neural networks for high-dimensional relationship mapping, a fully-connected network (FCN) can achieve acceptable accuracy for springback prediction [9]. Specifically, SP-NET and ES-NET each contain a hidden layer with 10 units. In addition, ES-NET also includes an implicit output layer (FCN2) for single-layer tube shape parameter equivalent, which is also used as an FCN with 2 units in the second stage.

### 3.2 Preliminary analysis of BMT equivalence section

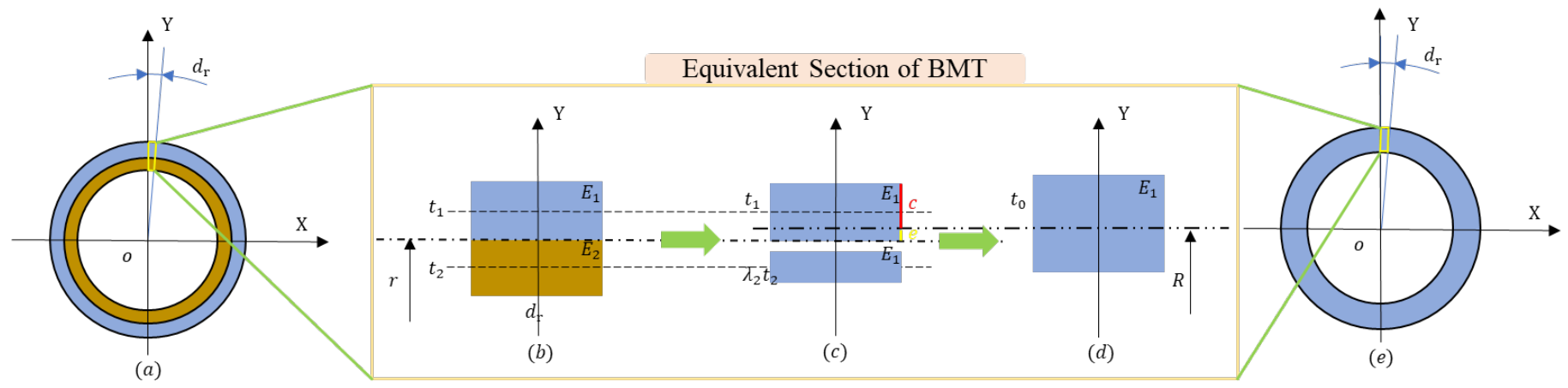

**Fig. 4.** Schematic diagram of equivalent section process: (a) is the BMT section; (b) is one of the microelements; (c) is the cross-section after the material $E_2$ is equivalent to $E_1$; (d) is the $E_1$ material section of the equivalent moment of inertia with (c); (e) is the equivalent single-layer tube section.

For the theory-driven pre-exploration of the ES-NET parameter in the first stage, the preliminary analysis based on equivalent section theory is necessary. As shown in Fig. 4, take the elastic modulus $E_1$ of material on the outside as the reference, based on the calculation of area moments, the distance between the central axis and the upper edge of the micro-element $c$ can be calculated. It can be seen from Eq. (5) that the area moments before and after are equal. Therefore, the radius $R$ of the equivalent section is

$$R = r + e = \mathrm{r} + \frac{(t_1+\sqrt{\lambda_2}t_2)(t_1-\sqrt{\lambda_2}t_2)}{2(t_1+\lambda_2 t_2)} \tag{6}$$

Where $r$ is the radius of the junction of the BMT section, $e$ is the centroid axis offset of the equivalent micro-element section compared with the original BMT, $t_i$ $(i = 1,2)$ is the thickness of the $i$-th material, $\lambda_2 = E_2/E_1$.

With the properties of the tube section and Eq. (6), we have

$$t_0^3 + 4R^2 t_0 - \frac{1}{R}[(r+t_1)^4 - (1-\lambda_2)r^4 - \lambda_2(r-t_2)^4] = 0 \tag{7}$$

Combined with conditions of real numbers and engineering, the equivalent material thickness $t_0$ in the can be obtained.

According to $R$ and $t_0$, the shape parameters of the equivalent single-layer tube can be obtained. Since the assumptions and simplifications are adopted, the error must exist. However, the above analysis ensures the equivalent result is under the basic physical laws, which can guarantee the physical rationality of the ES-NET pre-exploration in the first stage, and the fine-tuning of the PE-NET in the second stage.

### 3.3 Composition of the loss function

As shown in Fig. 3, in the first stage, the parameter updates for ES-NET and SP-NET are based on theory-driven and data-driven, respectively, as shown below.

$$L_p = MSE(ES-NET(x_s^b) - f_{ES-NET}(x_s^b)) \tag{8}$$

$$L_d = MSE(PE-NET(x^b) - Y_{SA}) \tag{9}$$

Where $x^b$ is the input of the BMT, $x_s^b$ is the shape parameter input of the $x^b$, $ES-NET(x_s^b)$ reflect the shape parameter of the single-layer tube, which is also the implicit output of ES-NET and the implicit input of the PE-NET in the second stage, donated as $x_s^s$. $Y_{SA}$ is the predicted springback which the output of SP-NET in the first stage and the output of the PE-NET in the second stage. $f_{ES-NET}$ represents the theory-driven knowledge for pre-exploration of the ES-NET in the first stage.

In the second stage, the PS-NET is fine-tuned with a combination of update-driven approaches. Based on Eq. (10) and Eq. (11), $L_p$ can keep the ES-NET in the reasonable parameter domain and guarantee the physical plausibility of this process. However, the reliability of the loss function $L_d$ for parameter update driven by real data will significantly be higher than $L_p$ driven by physical equivalence theory. In order to keep their advantages in the second training stage, the integration of the two parameter update-driven approaches for better cooperation is necessary. Therefore, the dynamic weight loss function will be adopted, as follows:

$$L_{ES-NET} = zL_p + (1-z)L_d \tag{10}$$

$$z = \begin{cases} P_{N(x_s^s,1)}(2x_s^s - f_{ES-NET}(x_s^b) \le X \le f_{ES-NET}(x_s^b)), & L_p > 2x_s^s \\ 0, & L_p \le 2x_s^s \end{cases} \tag{11}$$

Where $z$ is the dynamic weight coefficient and $N$ is the normal function.

Given the reasonable range of equivalence relationship result, in the second stage, when the implicit result of ES-NET is far away from the physical reasonable, $L_p$ will guide the parameters until it is physical reasonable. When the result is within the reasonable range, $L_p$ will no longer play a dominant role, and instead, the PE-NET parameter will be updated mainly based on $L_d$.

## 4 Case study

As one of the most commonly used material combinations, copper-aluminum BMT has received more and more attention due to its advantages in weight and thermal properties. Their material properties vary greatly, and their elastic modulus are $E_1$=80.7 GPa and $E_2$=110 Gpa, respectively. In engineering applications, compared with the widely used aluminum tube, the data of the copper-aluminum BMT is still rare, and the research on its deformation still needs to be improved, which makes it selected as the research object.

### 4.1 Dataset construction

FE simulation is an effective method to obtain engineering solutions [4, 18]. It is an important step to verify technical routes by constructing datasets that meet experimental requirements and reduce trial costs before actual applications.

With the Latin hypercube sampling, the datasets are determined. The ABAQUS 2016 platform is used for simulation analysis. The S4R shell with the specially 9th-order Simpson integration points in the thickness direction is used for the tube deformation unit, and the R3D4 rigid for the molds. The stress-strain trend of the material adopts the power hardening model. The friction constraint is applied when the tube blank is pulled. The composite material is assigned to the properties of the BMT, including the relative sampling thickness ratio. Since the velocity is one of the important factors affecting springback, the bending processing adopts explicit dynamics analysis, while static analysis is used for springback simulation. The simulated result is imported into MATLAB for post-processing to extract the springback result and build the dataset.

In order to meet the engineering practice, there is a significant order of magnitude difference between the single-layer tube and the BMT. Specifically, the single-layer tube dataset, namely Dataset1, contains 600 samples, provides sufficient single-layer tube deformation knowledge. Since the double-layer tube is a small sample, its dataset, i.e. Dataset2 has only 80 samples.

### 4.2 Precision analysis of proposed method

As mentioned in Sect. 3.1, the training consists of two stages. The first stage is for the parameter pre-exploration of the ES-NET and the pretrain of the SP-NET, and the second stage is for the parameter fine-tuning of PE-NET. Accuracy is the median error of multiple training results. The accuracy of the two stages is the median error of 30 training sets, as shown in Table 1. Particularly, in the second stage, PE-NET is constructed based on specific SP-NET and ES-NET with the median of error.

All training is performed on MATLAB 2022a on the same GPU. All datasets will be split before training, 80% of which will be used for training, and the remaining be used for tests. The test set does not participate in the training process. The Adam optimizer is employed, and the dataset is shuffled every epoch. For the first stage, the training minibatch size is set to 5, and the initial and decay factor of the gradient are 0.005 and 0.9, respectively. For the second stage, the training minibatch size is set to 2, and the initial and decay factor of the gradient are 0.0001 and 0.8, respectively.

**Table 1.** The median RMSE of the two training stages

| | SP-NET | ES-NET | |
|---|---|---|---|
| Stage 1 | Springback angle | $D_o$ | $T$ |
| | 0.7672 | 0.7133 | 0.4916 |
| | PE-NET | | |
| Stage 2 | Springback angle | | |
| | 0.3922 | | |

As shown in Table 1, the median error of PE-NET is 0.3922, which meets the requirements of engineering applications. It should be noted that although the second stage is fine-tuning based on the small-sample Dataset2, the accuracy is higher than that of SP-NET only based on data-driven. On the one hand, this reflects the rationality and effectiveness of the architecture proposed. On the other hand, based on the learning of multiple driving logic and datasets, the network can learn a variety of features and noise, which greatly improves the ability of generalization and overfit prevention. This can also be proved in Sect. 4.3.

## 4.3 Effectiveness of PE-NET

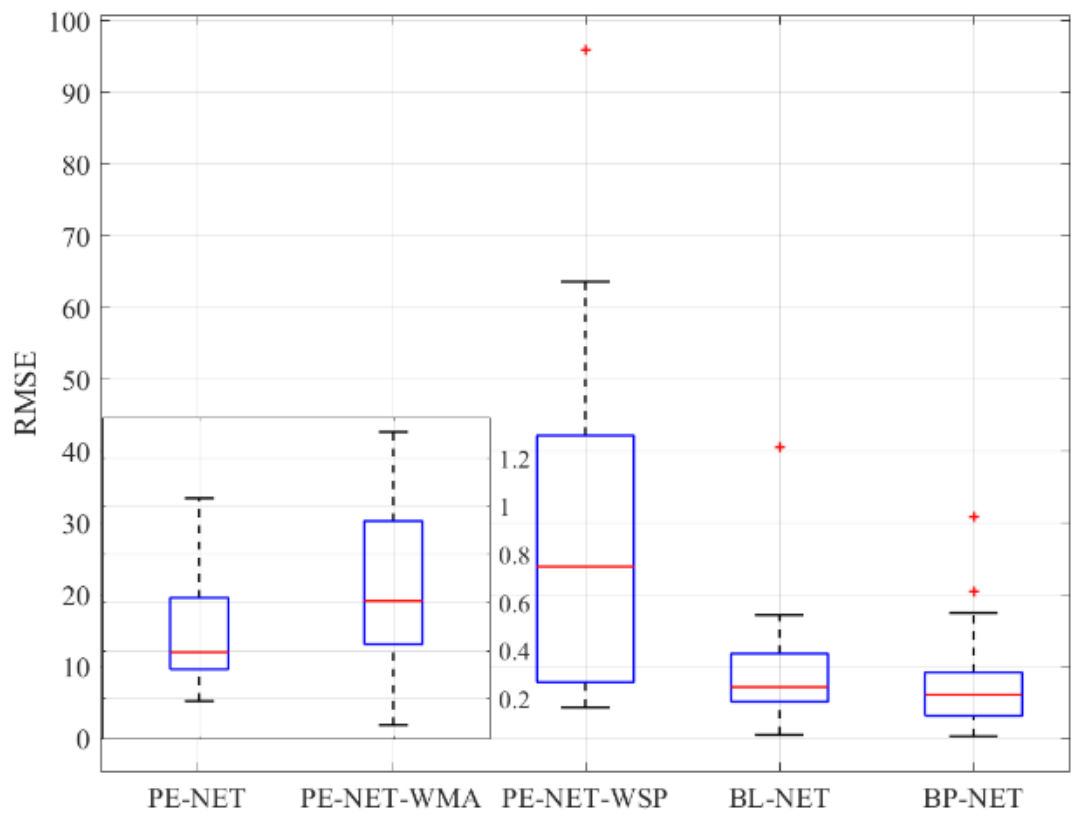

**Fig. 5.** Box-plot of springback prediction methods comparison

Controlled experiments are conducted to verify the effectiveness of the proposed physical logic architecture. PE-NET without ES-NET pre-exploration based on mechanism analysis, and without SP-NET pre-training are analyzed, denoted as PE-NET-WMA and PE-NET-WSP, respectively. Without any pre-operation, the PE-NET architecture and simple BPNN with 10 units hidden layer are trained based on Dataset2, denoted as BL-NET and BP-NET, respectively. All training parameters are the same. The results of multiple sets of training and corresponding average RMSE are recorded, as shown in Fig. 5 and Table 2, respectively.

**Table 2.** The median RMSE of the controlled experiments

| Method | RMSE |
|---|---|
| PE-NET | 0.3922 |
| PE-NET-WMA | 0.6019 |
| PE-NET-WSP | 23.7527 |
| BL-NET | 6.9424 |
| BP-NET | 6.3572 |

Results show that PE-NET performs the best no matter of accuracy or stability. Despite the possibility of available prediction accuracy, the lack of theoretical guidance and constraints reduces the stability of PE-NET-WMA, making it less efficient than PE-NET in engineering applications. Similarly, without pre-training with close domain knowledge, the performance of PE-NET-WSP is extremely poor, making it difficult to be applied to BMT springback prediction of small samples. The above two results also prove the validation of the deformation knowledge of single-layer tube and mechanism analysis, and the effectiveness of the proposed physical logic-based PE-NET architecture has also been demonstrated. For BL-NET and BP-NET, its poor accuracy and stability make it impossible to be used in engineering applications either.

In addition, the accuracy of the mechanism analysis is also revealed. The RMSE of equivalent Dataset2 springback is 1.6164. This also proves the necessity of fine-tuning based on BMT data. On this basis, it is feasible to further improve the equivalent theory with the help of the fine-tuned ES-NET, which also demonstrates the potential of the PE-NET architecture in interpretability.

## 5 Conclusion

In this work, we proposed a physical logic-based architecture network PE-NET to predict the springback of BMT with small samples. The BMT was logically equivalent to the single-layer tube, and then predicted with the single-layer springback knowledge. At first, the bending deformation mechanical analysis for section equivalent was conducted. Then, with the data-driven and theory-driven methods, the ES-NET and SP-NET were built with the theory-driven parameter pre-exploration and the data-driven pretraining, respectively. Finally, the PE-NET was constructed with the combination of ES-NET and SP-NET under the physical logic and the composition of the loss function. The validation and stability of the proposed method were verified with the FE simulation platform. This work is a primary attempt to solve the engineering problems with the only small-sample and limited theory, and will be integrated into more complex prediction and interpretability research in the future.

**Acknowledgments.**
This paper is funded by the Joint Funds of the National Natural Science Foundation of China (U20A20287), the National Natural Science Foundation of China (51905476), the Public Welfare Technology Application Projects of Zhejiang Province, China (LGG22E050008).